\documentclass{article}

\usepackage{arxiv}

\usepackage[utf8]{inputenc} 
\usepackage[T1]{fontenc}    
\usepackage{hyperref}       
\usepackage{url}            
\usepackage{booktabs}       
\usepackage{amsfonts}       
\usepackage{nicefrac}       
\usepackage{microtype}      
\usepackage{lipsum}

\usepackage{bm}
\usepackage{amsmath}
\usepackage{amssymb}
\usepackage{amsfonts}
\usepackage{graphicx}
\def\vec#1{\mathbf{#1}}
\usepackage{algorithm}
\usepackage{algorithmic}

\title{Meta-learning One-class Classifiers with Eigenvalue Solvers for Supervised Anomaly Detection}

\author{Tomoharu Iwata\\ 
  NTT Communication Science Laboratories, Kyoto, Japan
  \And Atsutoshi Kumagai\\
  NTT Software Innovation Center, Tokyo, Japan}
\date{}

\begin{document}
\maketitle

\begin{abstract}
Neural network-based anomaly detection methods have shown to achieve high performance. However, they require a large amount of training data for each task. We propose a neural network-based meta-learning method for supervised anomaly detection. The proposed method improves the anomaly detection performance on unseen tasks, which contains a few labeled normal and anomalous instances, by meta-training with various datasets. With a meta-learning framework, quick adaptation to each task and its effective backpropagation are important since the model is trained by the adaptation for each epoch. Our model enables them by formulating adaptation as a generalized eigenvalue problem with one-class classification; its global optimum solution is obtained, and the solver is differentiable. We experimentally demonstrate that the proposed method achieves better performance than existing anomaly detection and few-shot learning methods on various datasets.
\end{abstract}

\section{Introduction}

Anomaly detection is a task to find anomalous instances that
do not conform to expected behavior in a dataset~\cite{chandola2009anomaly}.
Anomaly detection has been used
in a wide variety of applications~\cite{patcha2007overview,hodge2004survey},
which include 
network intrusion detection~\cite{dokas2002data,yamanishi2004line},
fraud detection~\cite{aleskerov1997cardwatch},
defect detection~\cite{fujimaki2005approach,ide2004eigenspace},
and disease outbreak detection~\cite{wong2003bayesian}.
Deep learning-based anomaly methods
have shown to achieve high performance due to the high representation learning capability~\cite{hawkins2002outlier,ma2013parallel,sakurada2014anomaly,xu2015learning,erfani2016high,andrews2016detecting,lawson2017finding,schlegl2017unsupervised,chen2017outlier,akcay2018ganomaly,zenati2018efficient,ruff2018deep,ruff2020deep,chalapathy2019deep,schlegl2017unsupervised,sabokrou2018adversarially,ravanbakhsh2019training}. 
However, these methods require a large amount of training data.
Recently, few-shot learning, or meta-learning, methods attract attentions
for improving performance with a few labeled data~\cite{schmidhuber:1987:srl,bengio1991learning,ravi2016optimization,andrychowicz2016learning,vinyals2016matching,snell2017prototypical,bartunov2018few,finn2017model,li2017meta,kimbayesian,finn2018probabilistic,rusu2018meta,yao2019hierarchically,edwards2016towards,garnelo2018conditional,kim2019attentive,hewitt2018variational,bornschein2017variational,reed2017few,rezende2016one,tang2019,narwariya2020meta,xie2019meta,lake2019compositional}.
However, existing few-shot learning methods are not designed for anomaly detection.
The anomalous class is difficult to be explicitly modeled
since the definition of anomaly is that instances that are different from normal instances,
and there would various types of anomalies; some types of anomalies might not appear in training data.

In this paper, we propose a meta-learning method
for supervised anomaly detection.
The proposed model is meta-trained using various datasets,
such that it can improve the expected anomaly detection performance
on unseen tasks with a few labeled normal and anomalous instances.
We assume that normal, or non-anomalous, instances are located inside a hypersphere in a latent space,
and anomalous instances are located outside the hypersphere as
one-class classification-based anomaly detection methods
assume~\cite{moya1993one,scholkopf2001estimating,ruff2018deep,ruff2020deep}.
The one-class classification framework
has been successfully used for anomaly detection
since it does not explicitly model the anomaly class.
An anomaly score of an instance is calculated by the distance
between the center of the hypersphere and the instance in the latent space.

\begin{figure*}[t!]
  \centering
  \includegraphics[width=36em]{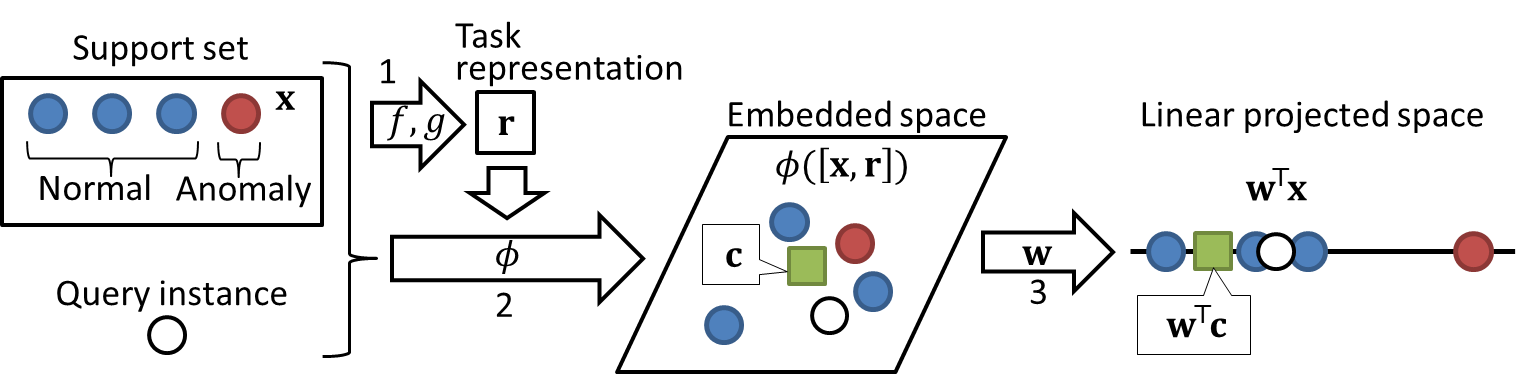}
  \caption{Our model. A blue (red) circle represents a normal (anomalous) instance in the support set,
    a white circle represents an unlabeled query instance, and a green square
    represents the center. 1) Task representation $\vec{r}$ is obtained using the support set by neural networks, $f$ and $g$. 2) Instances $\{\vec{x}\}$ in the original space are embedded
    into the embedded space by neural network $\phi$ using task representation $\vec{r}$.
    3) Embedded instances $\{\phi([\vec{x},\vec{r}])\}$ and center $\vec{c}$ are
    linearly projected into a one-dimensional space by $\vec{w}$ such that normal support instances
    are located close to the center while anomalous support instances are located far away from the center.
    Anomaly scores are calculated by the distance from the center in the linear projected space.
    All neural networks, $f$, $g$, and $\phi$, and center, $\vec{c}$, are shared across tasks, but linear projection, $\vec{w}$, is task-specific and estimated using the support set.}
  \label{fig:model}
\end{figure*}

A standard meta-learning framework learns how to adapt to small labeled data,
where the gradient on the adaptation is needed for each training epoch.
Therefore, it is important to quickly calculate
the adaptation to each task and its gradient.
We formulate the adaptation as a generalized eigenvalue problem
with one-class classification-based anomaly detection.
With the generalized eigenvalue problem,
we can find a global optimum solution,
and we can calculate the gradient of the solution in a closed form
using eigenvalues and eigenvectors~\cite{abou2009derivatives}.
Existing gradient-based meta-learning methods,
such as model-agnostic meta-learning~\cite{finn2017model},
calculate the adaptation to each task by iterative gradient descent steps.
They can only find a local optimum solution of the adaptation,
and they require the gradient of the iterative gradient descent steps,
which is computationally expensive.

Our model adapts to small labeled data,
which is called a support set,
by projecting instances into a task-specific latent space
with the following procedures
as illustrated in Figure~\ref{fig:model}.
First, a task representation is obtained using neural networks that take the support set as input.
Second, each instance is transformed into
a task-specific instance representation
by a neural network using the task representation.
With the high representation learning power of neural networks,
we can extract useful task-specific instance representations.
Third, the hypersphere center and
instance representations are linearly projected into a one-dimensional space
by a task-specific projection
to minimize the distance between the center and normal support instances
while maximizing the distance between the center and anomalous support instances.
The task-specific
projection is obtained by solving a generalized eigenvalue problem,
or in a closed form when the number of anomalous instances is one in the support set.
Since the optimum solution of the projection is differentiable,
we can backpropagate errors through the linear projection layer.
All the neural networks in our model are shared across tasks,
therefore our model is applicable to target tasks that are unseen in the training phase.

The neural networks in our model are trained by maximizing the expected test AUC
using an episodic training framework~\cite{ravi2016optimization,santoro2016meta,snell2017prototypical,finn2017model,li2019episodic}, where support sets and test instances are randomly generated
from training datasets to simulate target tasks for each epoch.
The AUC is the area under the receiver operating characteristic curve.
The AUC is commonly used for an evaluation measurement for anomaly detection~\cite{liu2012isolation}.

Our major contributes are summarized as follows:
1) We propose a few-shot learning method for anomaly detection.
2) We formulate adaptation with
  one-class classification-based anomaly detection
  as a generalized eigenvalue problem, 
  by which our model can be trained effectively
  to improve the expected test performance
  when adapted to small labeled data.
3) We experimentally demonstrate that the proposed method achieves better performance than existing anomaly detection and few-shot learning methods on various datasets.


\section{Related work}
\label{sec:related}

Our model is based on deep support vector data description (SVDD)~\cite{ruff2018deep}, and its supervised version~\cite{ruff2020deep},
which achieved better anomaly detection performance
than other deep learning-based methods.
The SVDD embeds instances into a latent space using a neural network
such that normal instances are located close to the center.
Our model extends the SVDD by incorporating support set information into the embeddings,
which enables us to detect anomalies with a few labeled instances for new target tasks.
Although some few-shot learning methods for anomaly detection have been
proposed~\cite{kruspe2019one,frikha2020few},
these methods assume that anomaly instances are not given in target tasks,
which is different from our setting.

Model-agnostic meta-learning (MAML)~\cite{finn2017model}
trains the model such that the performance is improved
when adapted to a support set by an iterative gradient descent method.
The backpropagation through the gradient descent steps is costly in terms of memory,
and thus the total number of steps must be kept small~\cite{abou2009derivatives}.
In contrast, our model adapts to a support set
by solving a generalized eigenvalue problem,
where a global optimum solution is obtained, and
the gradient of eigenvectors
is efficiently calculated using eigenvalues and eigenvectors
in a closed form~\cite{abou2009derivatives}.
Ridge regression differentiable discriminator~\cite{bertinetto2018meta}
and meta-learning with differentiable convex optimization~\cite{lee2019meta}
find a global optimum linear projection for quick adaptation in meta-learning,
although they consider classification tasks,
and solves a linear least square problem~\cite{bertinetto2018meta} or
a convex optimization problem~\cite{lee2019meta}.
On the other hand, we newly design a generalized eigenvalue problem
for quick adaptation in anomaly detection tasks.

Our model obtains a task representation
from a support set using neural networks,
which is similar to encoder-decoder style
meta-learning methods~\cite{xu2019metafun},
such as neural processes~\cite{garnelo2018conditional,garnelo2018neural},
where adaptation is performed by neural networks.
The encoder-decoder style methods
can adapt quickly by forwarding the support set
to neural networks.
However, it is difficult to approximate the adaptation to any support data
only by neural networks, 
and they usually require large training datasets.
In contrast, the proposed method adapts a linear projection
to the support data by directly solving an optimization problem.

Transfer anomaly detection methods have been proposed
to transfer knowledge in source tasks to target tasks~\cite{xiao2015robust,fujita2018one,andrews2016transfer,ide2018collaborative,ide2017multi}.
However, they require target data in a training phase.
Although there are transfer anomaly detection methods that do not use target data
for training~\cite{kumagai2019transfer,chen2014transfer},
these methods cannot use anomaly information in target data.

\section{Proposed method}
\label{sec:proposed}

\subsection{Problem formulation}

In a training phase, 
we are given labeled datasets in $T$ tasks,
$\{\mathcal{D}_{t}\}_{t=1}^{T}$,
where
$\mathcal{D}_{t}=\{(\vec{x}_{tn},y_{tn})\}_{n=1}^{N_{t}}$
is the $t$th task's dataset,
$\vec{x}_{tn}\in\mathbb{R}^{M}$ is the $n$th attribute vector,
and $y_{tn}\in\{0,1\}$ is its anomaly label, $y_{tn}=1$ if it is anomaly and $y_{tn}=0$ otherwise.
We assume that the attribute size $M$ is the same in all the training and target tasks.
In a test phase,
we are given a few labeled instances $\mathcal{S}=\{(\vec{x}_{n},y_{n})\}_{n=1}^{N_{\mathrm{S}}}$
in a target task, which is different from training tasks.
Our aim is to identity whether unlabeled instance $\vec{x}$ in the target task
is anomaly or not.
We call a few labeled instances $\mathcal{S}$ a support set, and call unlabeled instance $\vec{x}$ a query.


\subsection{Model}
\label{sec:model}

Our model outputs a task-dependent anomaly score of query $\vec{x}$
that is adapted to support set $\mathcal{S}$.
Figure~\ref{fig:model} illustrates our model.

First, task representation $\vec{r}$ is obtained with permutation invariant
neural networks~\cite{zaheer2017deep} taking support set $\mathcal{S}$ as input,
\begin{align}
  \vec{r} = g\Bigl(\frac{1}{N_{\mathrm{S}}}\sum_{(\vec{x},y)\in\mathcal{S}}f([\vec{x},y])\Bigr),
  \label{eq:r}
\end{align}
where $f$ and $g$ are feed-forward neural networks shared across tasks,
and $[\cdot,\cdot]$ is a concatenation.
Eq.~(\ref{eq:r}) outputs the same value even when instances in support set $\mathcal{S}$
are permutated since the summation is permutation invariant.
We use the permutation invariant neural network
since the order of the instances in the support set
should not affect the support set representation~\cite{garnelo2018conditional}.
Then, our model obtains a representation of instance $\vec{x}$ that depends on the support set
by nonlinearly transforming the concatenation of the instance attributes and support set representation,
$\phi([\vec{x},\vec{r}])\in\mathbb{R}^{J}$,
where $\phi$ is a feed-forward neural network shared across tasks.

Let $\vec{c}\in\mathbb{R}^{J}$ be a center
where normal instances are located close to
in the embedding space.
The center is shared across tasks.
Our model calculates an anomaly score of instance $\vec{x}$
by the distance between instance representation $\phi([\vec{x},\vec{r}])$ and
center $\vec{c}$
when they are linearly projected to a one-dimensional space as follows,
\begin{align}
  a(\vec{x}|\mathcal{S})=\parallel\hat{\vec{w}}^{\top}\phi([\vec{x},\vec{r}])-\hat{\vec{w}}^{\top}\vec{c}\parallel^{2},
  \label{eq:a}
\end{align}
where $\hat{\vec{w}}\in\mathbb{R}^{J}$ is a task-specific
linear projection vector.
Instances that are located far away from $\vec{c}$ have high anomaly scores,
and instances that are located close to $\vec{c}$ have low anomaly scores.
We adapt $\hat{\vec{w}}$ to support set $\mathcal{S}$
but do not adapt the other parameters,
i.e., parameters of neural networks $f$, $g$, $\phi$, and center $\vec{c}$.
Then, the global optimum solution for adaptation
is obtained in the following way without iterative gradient descent steps.
We adapt projection vector $\hat{\vec{w}}$ to support set $\mathcal{S}$
by maximizing anomaly scores of anomalous support instances
while minimizing anomaly scores of normal support instances,
which is achieved by solving the following optimization problem,
\begin{align}
  \hat{\vec{w}}=
  \arg\max_{\vec{w}}\frac{\frac{1}{N_{\mathrm{A}}}\sum_{\vec{x}\in\mathcal{S}_{\mathrm{A}}}a(\vec{x}|\mathcal{S})}{\frac{1}{N_{\mathrm{N}}}\sum_{\vec{x}\in\mathcal{S}_{\mathrm{N}}}a(\vec{x}|\mathcal{S})+\eta\parallel\vec{w}\parallel^{2}}
  = \arg\max_{\vec{w}}\frac{\mathrm{tr}(\vec{w}^{\top}\vec{S}_{\mathrm{A}}\vec{w})}{\mathrm{tr}(\vec{w}^{\top}\vec{S}_{\mathrm{N}}\vec{w})},
  \label{eq:w}
\end{align}
where
$\mathcal{S}_{\mathrm{A}}=\{\vec{x}|y=1,(\vec{x},y)\in\mathcal{S}\}$ is the set of anomalous support instances,
$N_{\mathrm{A}}$ is its size,
$\mathcal{S}_{\mathrm{N}}=\{\vec{x}|y=0,(\vec{x},y)\in\mathcal{S}\}$ is the set of normal support instances,
$N_{\mathrm{N}}$ is its size, $\eta>0$ is a parameter to be trained, and
\begin{align}
  \vec{S}_{\mathrm{A}}=\frac{1}{N_{\mathrm{A}}}\sum_{\vec{x}\in\mathcal{S}_{\mathrm{A}}}(\phi([\vec{x},\vec{r}])-\vec{c})(\phi([\vec{x},\vec{r}])-\vec{c})^{\top},\\
  \vec{S}_{\mathrm{N}}=\frac{1}{N_{\mathrm{N}}}\sum_{\vec{x}\in\mathcal{S}_{\mathrm{N}}}(\phi([\vec{x},\vec{r}])-\vec{c})(\phi([\vec{x},\vec{r}])-\vec{c})^{\top}+\eta\vec{I}.
\end{align}
The numerator of Eq.~(\ref{eq:w}) is the mean of anomaly scores of anomalous support instances,
and the denominator is the mean of anomaly scores of normal support instances
with regularization term $\eta\parallel\vec{w}\parallel^{2}$.
With the regularization term, $\vec{S}_{\mathrm{N}}$ is positive definite,
and the optimization becomes stable.
Note that Eq.~(\ref{eq:w}) is different
from the Fisher linear discriminant analysis (FLDA).
FLDA maximizes the between-class covariance
while minimizing the within-class covariance.
In contrast,
our model maximizes the distance between anomalous instances and the center
while minimizing the distance between normal instances and the center.


Since Eq.~(\ref{eq:w}) is a generalized Rayleigh quotient~\cite{parlett1998symmetric},
we can obtain its solution by
solving the following generalized eigenvalue problem
in a similar way to FLDA~\cite{welling2005fisher,ghojogh2019fisher},
\begin{align}
  \vec{S}_{\mathrm{A}}\hat{\vec{w}}=\lambda\vec{S}_{\mathrm{N}}\hat{\vec{w}},
  \label{eq:eigen}
\end{align}
where $\lambda$ is the largest eigenvalue, and $\hat{\vec{w}}$ is its corresponding eigenvector.
The eigenvector of the generalized eigenvalue problem is differentiable
since it can be solved with eigenvalue decomposition and 
matrix inverse~\cite{welling2005fisher}, which are differentiable.
Therefore, we can backpropagate errors through the linear projection
based on the generalized eigenvalue problem.
The gradient of eigenvectors
is efficiently calculated using eigenvalues and eigenvectors
in a closed form~\cite{abou2009derivatives}.
When the dimensionality of the linearly projected space is more than one,
optimization problem Eq.~(\ref{eq:w}) is not
a generalized Rayleigh quotient,
and the solution cannot be given
via a generalized eigenvalue problem~\cite{yan2006trace,cunningham2015linear}.
Therefore, instances are projected in a one-dimensional space.
When the number of anomalous instances is one,
there exists a simpler solution for Eq.~(\ref{eq:w}),
\begin{align}
  \hat{\vec{w}}\propto\vec{S}_{\mathrm{N}}^{-1}(\phi([\vec{x}_{\mathrm{A}},\vec{r}])-\vec{c}),
  \label{eq:w1}
\end{align}
where $\vec{x}_{\mathrm{A}}$ is the anomalous support instance.
The derivations of Eqs.~(\ref{eq:w},\ref{eq:w1})
are described in Appendices~\ref{app:eq3} and \ref{app:eq7}.

\subsection{Training}

In our model, parameters to be estimated $\bm{\Theta}$ are
the parameters of neural networks $f$, $g$, and $\phi$, and
regularization parameter $\eta$,
all of which are shared across different tasks.
We determine center $\vec{c}$ as in SVDD~\cite{ruff2018deep},
where after initializing the neural network parameters, we fix the center
by the mean of the embeddings of the training normal instances.
Note that we do not need to train task-specific linear projection vector $\hat{\vec{w}}$
by gradient-based methods since it is calculated by solving a generalized eigenvalue problem
as shown in Section~\ref{sec:model}.

We train our model by maximizing the expected test AUC.
Let $\mathcal{Q}$ be a query set, which is a set of query instances,
and $a(\cdot|\mathcal{S})$ be the anomaly score function adapted by support set $\mathcal{S}$.
The empirical AUC is given by the probability that 
scores of anomalous instances are higher than those of normal instances.
The empirical AUC of query set $\mathcal{Q}$ with anomaly score function $a(\cdot|\mathcal{S})$ is calculated by
\begin{align}
  \mathrm{AUC}(\mathcal{Q}|a(\cdot|\mathcal{S}))
  =\frac{1}{N^{\mathrm{Q}}_{\mathrm{A}}N^{\mathrm{Q}}_{\mathrm{N}}}
  \sum_{\vec{x}\in\mathcal{Q}_{\mathrm{A}}}\sum_{\vec{x'}\in\mathcal{Q}_{\mathrm{N}}}I(a(\vec{x}|\mathcal{S})>a(\vec{x}'|\mathcal{S})),
  \label{eq:auc}
\end{align}
where $I$ is the indicator function,
$I(A)=1$ if $A$ is true, $I(A)=0$ otherwise,
$\mathcal{Q}_{\mathrm{A}}$ is a set of anomalous query instances,
$N_{\mathrm{A}}^{\mathrm{Q}}$ is its size,
$\mathcal{Q}_{\mathrm{N}}$ is a set of normal query instances,
and $N_{\mathrm{N}}^{\mathrm{Q}}$ is its size.
The indicator function is not differentiable.
To make the empirical AUC differentiable,
we use sigmoid function $\sigma(a(\vec{x}|\mathcal{S})-a(\vec{x}'|\mathcal{S}))=\frac{1}{1+\exp(-(a(\vec{x}|\mathcal{S})-a(\vec{x}'|\mathcal{S})))}$
instead of indicator function $I(a(\vec{x}|\mathcal{S}) > a(\vec{x}'|\mathcal{S}))$, which is often used for a smooth approximation of the indicator function~\cite{ma2005regularized}.
Let 
\begin{align}
  \widetilde{\mathrm{AUC}}(\mathcal{Q}|a(\cdot|\mathcal{S}))
  =\frac{1}{N^{\mathrm{Q}}_{\mathrm{A}}N^{\mathrm{Q}}_{\mathrm{N}}}
  \sum_{\vec{x}\in\mathcal{Q}_{\mathrm{A}}}\sum_{\vec{x'}\in\mathcal{Q}_{\mathrm{N}}}
  \sigma(a(\vec{x}|\mathcal{S})-a(\vec{x}'|\mathcal{S})),
  \label{eq:auc_smooth}
\end{align}
be a smoothed version
of $\mathrm{AUC}(\mathcal{Q}|a(\cdot|\mathcal{S}))$.
The objective function to be maximized is the expected smoothed test empirical AUC
as follows,
\begin{align}
  \mathbb{E}_{t\sim\{1,\cdots,T\}}[\mathbb{E}_{(\mathcal{S},\mathcal{Q})\sim\mathcal{D}_{t}}[\widetilde{\mathrm{AUC}}(\mathcal{Q}|a(\cdot|\mathcal{S}))]],
  \label{eq:E}
\end{align}
where $\mathbb{E}$ represents an expectation.
The objective function is maximized using an episodic training
framework~\cite{ravi2016optimization,santoro2016meta,snell2017prototypical,finn2017model,li2019episodic},
where support and query sets are randomly generated
from training datasets to simulate target tasks for each epoch.
Algorithm~\ref{alg} shows the training procedure for our model.
In Line~2,
we fix center $\vec{c}$ as in SVDD~\cite{ruff2018deep}
by the mean of the embeddings of the training normal instances
using the neural networks with the initial model parameters.
In Lines~4--6, we generate support and query set from training datasets
to simulate a target task.
We obtain the projection vector using the support set in Line~7.
The loss, which is the negative smoothed AUC, and its gradients are calculated in Line~8.
In Line~9, we update the model parameters with a stochastic gradient-based method,
such as Adam~\cite{kingma2014adam}.

The computational complexity of each training step for each task is:
$O(N_{\mathrm{S}})$ for obtaining task representation $\vec{r}$ using the support set
in Eq.~(\ref{eq:r}),
$O(N_{\mathrm{S}}+N_{\mathrm{S}}+N_{\mathrm{A}}^{\mathrm{Q}}+N_{\mathrm{N}}^{\mathrm{Q}})$ for
calculating instance representations $\phi([\vec{x},\vec{r}])$ for all instances in
the support and query sets,
$O(J^{3})$ for solving the generalized eigenvalue problem in Eq.~(\ref{eq:w}),
and $O(N_{\mathrm{A}}^{\mathrm{Q}}N_{\mathrm{N}}^{\mathrm{Q}})$ for evaluating the empirical AUC in Eq.~(\ref{eq:auc}).
Although $O(J^{3})$ is a demanding part,
since $J$ is the output layer size of neural network $\phi$, we can control $J$.

\begin{algorithm}[t!]
  \caption{Training procedure of our model. $\mathrm{RandomSample}(\mathcal{D},N)$ generates a set of $N$ elements chosen uniformly at random from set $\mathcal{D}$ without replacement.}
  \label{alg}
  \begin{algorithmic}[1]
    \renewcommand{\algorithmicrequire}{\textbf{Input:}}
    \renewcommand{\algorithmicensure}{\textbf{Output:}}
    \REQUIRE{Training datasets $\{\mathcal{D}_{t}\}_{t=1}^{T}$,
      number of support instances $N_{\mathrm{S}}$, number of query instances $N_{\mathrm{Q}}$}
    \ENSURE{Trained model parameters $\bm{\Theta}$}
    \STATE Initialize model parameters $\bm{\Theta}$
    \STATE Fix center $\vec{c}$
    \WHILE{End condition is satisfied}
    \STATE Randomly select task index,\\$t\gets\mathrm{RandomSample}(\{1,\cdots,T\},1)$
    \STATE Randomly generate support set,\\$\mathcal{S}\gets\mathrm{RandomSample}(\mathcal{D}_{t},N_{\mathrm{S}})$
    \STATE Randomly generate query set,\\$\mathcal{Q}\gets\mathrm{RandomSample}(\mathcal{D}_{t}\setminus\mathcal{S},N_{\mathrm{Q}})$    
    \STATE Adapt anomaly score function $a(\cdot|\mathcal{S})$ to support set $\mathcal{S}$
    by solving a generalized eigenvalue problem in Eq.~(\ref{eq:eigen})
    \STATE Using the adapted anomaly score function, calculate loss $-\widetilde{\mathrm{AUC}}(\mathcal{Q}|a(\cdot|\mathcal{S}))$ and its gradients
    \STATE Update model parameters $\bm{\Theta}$ using the loss and its gradient
    \ENDWHILE
  \end{algorithmic}
\end{algorithm}

\subsection{Only normal instances in support sets}
\label{sec:normal}

In the previous subsections, we assume that both anomalous 
and normal instances are given in support sets.
When only normal instances are given in support sets,
we modify anomaly scores in Eq.~(\ref{eq:a})
by the distance between projected instance representation $\hat{\vec{w}}^{\top}\phi([\vec{x},\vec{r}])$ and
center $\vec{c}\in\mathbb{R}^{K}$
as follows,
$a(\vec{x}|\mathcal{S})=\parallel\hat{\vec{W}}^{\top}\phi([\vec{x},\vec{r}])-\vec{c}\parallel^{2}$,
where center $\vec{c}$ lies in the projected space,
and $\hat{\vec{W}}\in\mathbb{R}^{J\times K}$ is the projection matrix.
Projection matrix $\hat{\vec{W}}$ is obtained by
minimizing anomaly scores of normal support instances as follows,
$\hat{\vec{W}}=\arg\min_{\vec{W}}\frac{1}{N_{\mathrm{N}}}\sum_{\vec{x}\in\mathcal{S}_{\mathrm{N}}}a(\vec{x}|\mathcal{S})$.
With the least squares method, its global optimum solution is given in a closed form as follows,
$\hat{\vec{W}}=(\bm{\Phi}^{\top}\bm{\Phi})^{-1}\bm{\Phi}^{\top}\vec{C}$,
where $\bm{\Phi}=[\phi([\vec{x},\vec{r}])]_{\vec{x}\in\mathcal{S}_{\mathrm{N}}}$
is an ($N_{\mathrm{N}}\times J$) matrix of instance representations,
and $\vec{C}=[\vec{c},\cdots,\vec{c}]$ is an ($N_{\mathrm{N}}\times K$) matrix of the center.

\section{Experiments}
\label{sec:experiments}

\subsection{Data}

We evaluated the proposed method using 22 datasets for outlier detection in
\cite{campos2016evaluation}~\footnote{The outlier detection datasets were obtained from \url{http://www.dbs.ifi.lmu.de/research/outlier-evaluation/DAMI/}.},
Landmine dataset~\cite{xue2007multi},
and IoT datasets~\cite{meidan2018n,mirsky2018kitsune}~\footnote{The IoT datasets were obtained from \url{https://archive.ics.uci.edu/ml/datasets/detection_of_IoT_botnet_attacks_N_BaIoT}.}.
Each of the 22 outlier detection datasets contained data for a single outlier detection task.
We synthesized multiple tasks by multiplying a task-specific random matrix to attribute vectors,
where each element of the random matrix was generated from uniform randomly with range $[-1,1]$,
and the size of the random matrix was ($M\times M$), where
$M$ is the number of attributes of the dataset.
We generated 400 training, 50 validation, and 50 target tasks for each dataset.
The Landmine dataset contained attribute vectors extracted from radar images
and corresponding binary labels indicating whether landmine (anomaly) or clutter (normal)
at 29 landmine fields (tasks).
We randomly split the 29 tasks into 23 training, two validation, and four target tasks.
The IoT datasets consisted of traffic data
on nine IoT devices infected by ten types of attacks (anomaly) as well as benign (normal) traffic data.
We regarded an attack type on a device as a task, where there were 80 tasks in total.
We used tasks from a device for target,
90\% of the other tasks for training,
and the remaining tasks for validation.
The statistics for each dataset are described in Appendix~\ref{app:data}.
In all datasets, 
the number of normal support instances was $N_{\mathrm{N}}=5$,
the number of anomalous support instances was $N_{\mathrm{A}}=1$,
the number of normal query instances was $N^{\mathrm{Q}}_{\mathrm{N}}=25$,
and 
the number of anomalous query instances was $N^{\mathrm{Q}}_{\mathrm{A}}=5$.
We generated ten different splits of training, validation, and target tasks
for each dataset, and evaluated by the test AUC on target tasks averaged over the ten splits.

\subsection{Comparing methods}

We compared the proposed method
with the following 12 methods:
SVDD, SSVDD, Proto, R2D2, MAML, FT, OSVM, IF, LOF, LR, KNN, and RF.

SVDD was deep support vector data description~\cite{ruff2018deep},
and SSVDD was its supervised version.
With SVDD, a neural network was trained
so as to minimize the distance
between the center and normal support instances.
With SSVDD, the neural network was trained so as to maximize the AUC,
where anomaly scores were calculated by the distance from the center
as in the proposed method.
Proto was prototypical networks~\cite{snell2017prototypical},
which was a representative few-shot learning method for classification.
With Proto, attribute vectors were encoded by a neural network,
and class probabilities were calculated using the negative distance from 
the mean of the encoded support instances for each class.
R2D2 was ridge regression differentiable discriminators~\cite{bertinetto2018meta}.
R2D2 was a few-shot classification method based on a neural network,
where its final linear layer was determined
by solving a least square problem on a support set.
FT was a fine-tuning version of SSVDD,
where the neural network was fine-tuned using support sets of target tasks
after SSVDD was trained using training datasets.
MAML was model-agnostic meta-learning~\cite{finn2017model} of SSVDD,
where initial parameters of the neural network were trained so as to
maximize the expected test AUC when fine-tuned using a support set.
With SVDD, SSVDD, Proto, R2D2, and MAML,
we used an episodic training framework.
For the objective function, we used the expected test AUC with SSVDD, Proto, R2D2, FT,
and MAML as in the proposed method.

OSVM was one-class support vector machines~\cite{scholkopf2001estimating},
IF was isolation forest~\cite{liu2008isolation}, and
LOF was local outlier factors~\cite{breunig2000lof},
which were unsupervised anomaly detection methods.
LR was logistic regression,
KNN was the $k$-nearest neighbor method,
and RF was random forest~\cite{breiman2001random},
which were supervised classification methods.

The proposed method,
Proto, R2D2, MAML, and FT were meta-learning methods,
where they used training datasets as well as target support sets for calculating anomaly scores
of target query sets.
With SVDD and SSVDD, training datasets were used for training, but target support sets were not used.
OSVM, IF, LOF, LR, KNN, and RF were trained
using only the support set of target tasks.

\subsection{Implementation}

With the proposed method,
we used three-layered feed-forward neural networks for $f$ and $g$,
and four-layered feed-forward neural networks for $\phi$, where
their hidden and output layers contained 256 units.
For neural networks with SVDD, SSVDD, Proto, R2D2, FT, MAML,
we used four-layered feed-forward neural networks with
256 hidden and output units,
which were the same with neural network $\phi$ with the proposed method.
With SVDD and SSVDD as well as $\phi$ in the proposed method,
all the bias terms from the neural networks were removed
to prevent a hypersphere collapse~\cite{ruff2018deep}.
We optimized SVDD, SSVDD, Proto, R2D2, MAML, and the proposed method
using Adam~\cite{kingma2014adam} with learning rate $10^{-3}$,
dropout rate~\cite{srivastava2014dropout} $0.1$, and batch size 256.
The validation data were used for early stopping,
and the maximum number of epochs was 1,000.
The number of fine-tuning epochs with
FT was 30, and that with MAML was five.
We implemented the proposed method, SVDD, SSVDD, Proto, R2D2, MAML, and FT
with PyTorch~\cite{paszke2017automatic}.
For MAML, we used Higher, which
is a library for higher-order optimization~\cite{grefenstette2019generalized}.
We implemented OSVM, IF, LOF, LR, KNN, and RF with scikit-learn~\cite{pedregosa2011scikit},
and used their default parameters except for KNN,
where the number of neighbors was set to one.

\subsection{Results}

Table~\ref{tab:auc} shows the test AUC on target tasks for each dataset.
The proposed method achieved the best average performance.
SVDD and SSVDD calculated anomaly scores that were independent of tasks.
Therefore, their performance was low compared with the proposed method,
which calculated task-dependent anomaly scores.
With Proto, instances were embedded in a latent space that was the same for all tasks.
Therefore, Proto could not encode characteristics for each task in the embeddings.
On the other hand, the proposed method embedded instances in a task-specific latent space,
which resulted in the better performance of the proposed method than Proto.
Although R2D2 calculated task-specific anomaly scores,
it considered anomaly detection as a standard classification problem.
In contrast, the proposed method was designed for anomaly detection
based on one one-class classification-based methods~\cite{moya1993one,scholkopf2001estimating,ruff2018deep,ruff2020deep}, which have achieved good performance for anomaly detection.
MAML failed to improve the performance with these datasets
since it adapted the model with a small number of gradient steps.
Although FT improved the performance from SSVDD by fine-tuning, the test AUC was worse
than the proposed method.
It is because FT trains neural networks in two separate steps: pretraining and finetuning.
On the other hand, the proposed method
trained neural networks such that the expected test performance was improved
when optimized with small labeled data.
Since OSVM, IF, LOF, LR, KNN, and RF used only target support sets for training,
and could not exploit knowledge in training datasets, their test AUC was low.

\begin{table*}[t!]
  \centering
  \caption{Test AUC on target tasks for each dataset. Values in bold typeface are not statistically different at 5\% level from the best performing method in each dataset according to a paired t-test. The second bottom row shows the average test AUC over all datasets, and column shows the average AUC over
    all datasets, and the value in bold indicates the best average test AUC. The bottom row shows the number of datasets the method achieved the best or not statistically different from the best.}
  \label{tab:auc}
  {\tabcolsep=0.3em
    \begin{tabular}{lrrrrrrrrrrrrr}
      \hline
      & Ours & {\scriptsize SVDD} & {\tiny SSVDD} & Proto & {\scriptsize R2D2} & {\tiny MAML} & FT & {\tiny OSVM} & IF & LOF & LR & KNN & RF \\     
      \hline
ALOI & 0.999 & 0.920 & 0.983 & {\bf 0.999} & 0.995 & 0.290 & 0.650 & 0.781 & 0.781 & 0.833 & 0.947 & 0.531 & 0.876\\
Annthyroid & {\bf 0.739} & 0.573 & 0.549 & 0.643 & 0.653 & 0.463 & 0.531 & 0.401 & 0.400 & 0.179 & 0.403 & 0.200 & 0.343\\
Arrhythmia & 0.820 & 0.813 & {\bf 0.829} & 0.825 & 0.764 & 0.703 & 0.757 & 0.789 & 0.746 & 0.818 & 0.604 & 0.080 & 0.504\\
Cardiotoco. & {\bf 0.988} & 0.537 & 0.484 & 0.526 & 0.884 & 0.504 & 0.277 & 0.953 & 0.849 & 0.965 & 0.919 & 0.177 & 0.492\\
Glass & {\bf 0.952} & 0.682 & 0.657 & 0.879 & 0.894 & 0.649 & 0.719 & 0.836 & 0.802 & 0.253 & 0.881 & 0.473 & 0.674\\
HeartDisease & {\bf 0.715} & 0.582 & 0.580 & 0.592 & 0.603 & 0.544 & 0.646 & 0.487 & 0.414 & 0.328 & 0.474 & 0.237 & 0.417\\
Hepatitis & {\bf 0.843} & 0.660 & 0.687 & 0.818 & 0.819 & 0.575 & 0.801 & 0.797 & 0.584 & 0.721 & 0.826 & 0.456 & 0.648\\
InternetAds & 0.788 & 0.759 & 0.795 & 0.739 & 0.693 & 0.721 & 0.740 & {\bf 0.802} & 0.740 & 0.617 & 0.732 & 0.000 & 0.491\\
Ionosphere & 0.925 & 0.715 & 0.726 & 0.913 & 0.869 & 0.645 & {\bf 0.949} & 0.894 & 0.693 & 0.882 & 0.855 & 0.293 & 0.766\\
KDDCup99 & {\bf 0.976} & 0.967 & {\bf 0.975} & 0.965 & 0.907 & 0.882 & 0.969 & 0.975 & 0.849 & 0.803 & 0.642 & 0.676 & 0.566\\
Lympho. & {\bf 0.943} & 0.887 & 0.937 & 0.932 & 0.769 & 0.687 & 0.929 & 0.882 & 0.668 & 0.798 & 0.705 & 0.356 & 0.625\\
PageBlocks & 0.915 & 0.897 & 0.899 & 0.888 & 0.933 & 0.867 & 0.895 & {\bf 0.950} & 0.780 & 0.915 & 0.748 & 0.198 & 0.669\\
Parkinson & 0.884 & 0.776 & 0.699 & 0.885 & 0.884 & 0.607 & 0.856 & 0.877 & 0.647 & 0.875 & {\bf 0.910} & 0.588 & 0.715\\
PenDigits & {\bf 0.974} & 0.427 & 0.511 & 0.953 & 0.968 & 0.475 & 0.947 & 0.621 & 0.467 & 0.462 & 0.856 & 0.785 & 0.820\\
Pima & {\bf 0.730} & 0.701 & {\bf 0.725} & 0.488 & 0.603 & 0.662 & 0.599 & 0.704 & 0.650 & 0.557 & 0.548 & 0.122 & 0.398\\
Shuttle & 0.921 & 0.507 & 0.647 & 0.885 & {\bf 0.999} & 0.605 & 0.827 & 0.818 & 0.600 & 0.225 & 0.796 & 0.722 & 0.752\\
SpamBase & 0.733 & 0.639 & 0.607 & 0.752 & {\bf 0.774} & 0.456 & 0.672 & 0.414 & 0.377 & 0.073 & 0.659 & 0.276 & 0.446\\
Stamps & 0.946 & 0.871 & 0.897 & 0.828 & 0.930 & 0.797 & 0.935 & {\bf 0.966} & 0.865 & 0.933 & 0.847 & 0.477 & 0.762\\
WBC & {\bf 0.997} & {\bf 0.998} & 0.997 & 0.990 & 0.947 & 0.988 & 0.993 & 0.992 & 0.991 & 0.913 & 0.108 & 0.188 & 0.836\\
WDBC & {\bf 0.941} & 0.753 & 0.800 & 0.866 & 0.936 & 0.732 & 0.802 & 0.835 & 0.804 & 0.858 & 0.915 & 0.433 & 0.673\\
Waveform & {\bf 0.782} & 0.607 & 0.643 & 0.764 & 0.740 & 0.562 & 0.760 & 0.648 & 0.609 & 0.395 & 0.712 & 0.281 & 0.566\\
Wilt & {\bf 0.829} & 0.481 & 0.599 & 0.787 & 0.689 & 0.443 & 0.641 & 0.571 & 0.510 & 0.497 & 0.645 & 0.382 & 0.538\\
Landmine & {\bf 0.922} & 0.749 & {\bf 0.898} & 0.603 & 0.821 & {\bf 0.876} & 0.829 & 0.834 & 0.693 & 0.650 & 0.745 & 0.217 & 0.574\\
Doorbell1 & {\bf 0.997} & 0.904 & {\bf 0.996} & {\bf 0.961} & 0.945 & 0.979 & {\bf 0.997} & 0.934 & 0.736 & 0.771 & {\bf 0.989} & 0.837 & 0.972\\
Thermostat & {\bf 0.992} & 0.854 & 0.998 & {\bf 0.962} & {\bf 0.980} & 0.952 & {\bf 0.999} & 0.903 & 0.658 & 0.804 & 0.947 & 0.708 & 0.932\\
Doorbell2 & {\bf 0.990} & 0.795 & {\bf 0.995} & 0.972 & {\bf 0.954} & 0.949 & {\bf 0.991} & 0.868 & 0.688 & 0.461 & 0.934 & 0.504 & 0.740\\
Monitor & {\bf 0.996} & 0.859 & {\bf 0.995} & 0.970 & 0.977 & 0.966 & {\bf 0.998} & 0.919 & 0.737 & 0.690 & 0.972 & 0.842 & 0.983\\
Camera1 & {\bf 0.993} & 0.821 & {\bf 0.991} & 0.949 & 0.962 & 0.953 & {\bf 0.993} & 0.880 & 0.622 & 0.664 & 0.951 & 0.631 & 0.874\\
Camera2 & {\bf 0.995} & 0.802 & {\bf 0.995} & 0.983 & 0.980 & 0.902 & {\bf 0.995} & 0.844 & 0.598 & 0.651 & 0.945 & 0.810 & 0.958\\
Webcam & {\bf 0.998} & 0.833 & {\bf 0.999} & 0.987 & {\bf 0.990} & 0.945 & {\bf 0.999} & 0.923 & 0.674 & 0.826 & 0.971 & 0.804 & 0.983\\
Camera3 & {\bf 0.998} & 0.806 & {\bf 0.988} & 0.968 & 0.970 & 0.937 & {\bf 0.996} & 0.908 & 0.640 & 0.706 & 0.970 & 0.800 & 0.966\\
Camera4 & {\bf 0.998} & 0.872 & 0.962 & 0.948 & 0.932 & 0.951 & {\bf 0.992} & 0.959 & 0.753 & 0.808 & 0.977 & 0.611 & 0.964\\
\hline
Average & {\bf 0.913} & 0.751 & 0.814 & 0.851 & 0.868 & 0.727 & 0.834 & 0.811 & 0.676 & 0.654 & 0.785 & 0.459 & 0.704 \\
\hline
\#Best & {\bf 23} & 1 & 11 & 3 & 5 & 1 & 10 & 3 & 0 & 0 & 2 & 0 & 0 \\
\hline
  \end{tabular}}
\end{table*}

\begin{figure*}[t!]
  \centering
  \includegraphics[width=40em]{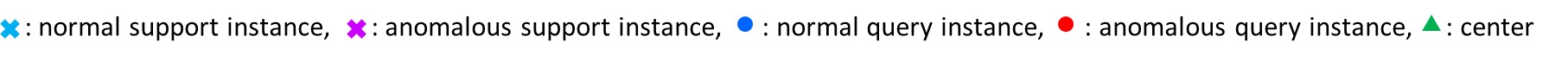}
  \\
  {\tabcolsep=-0.7em
  \begin{tabular}{ccc}
    \includegraphics[width=16em]{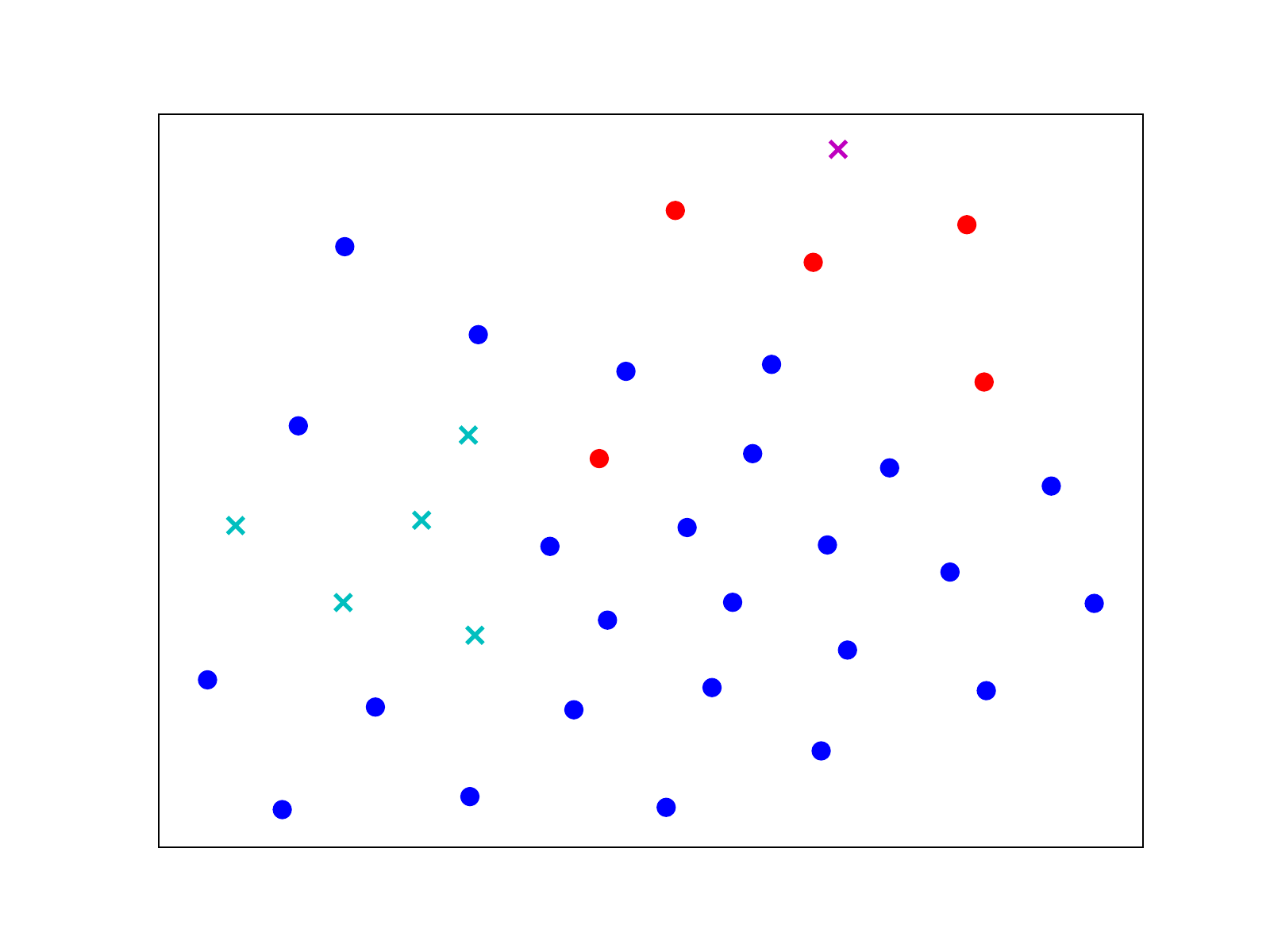}&
    \includegraphics[width=16em]{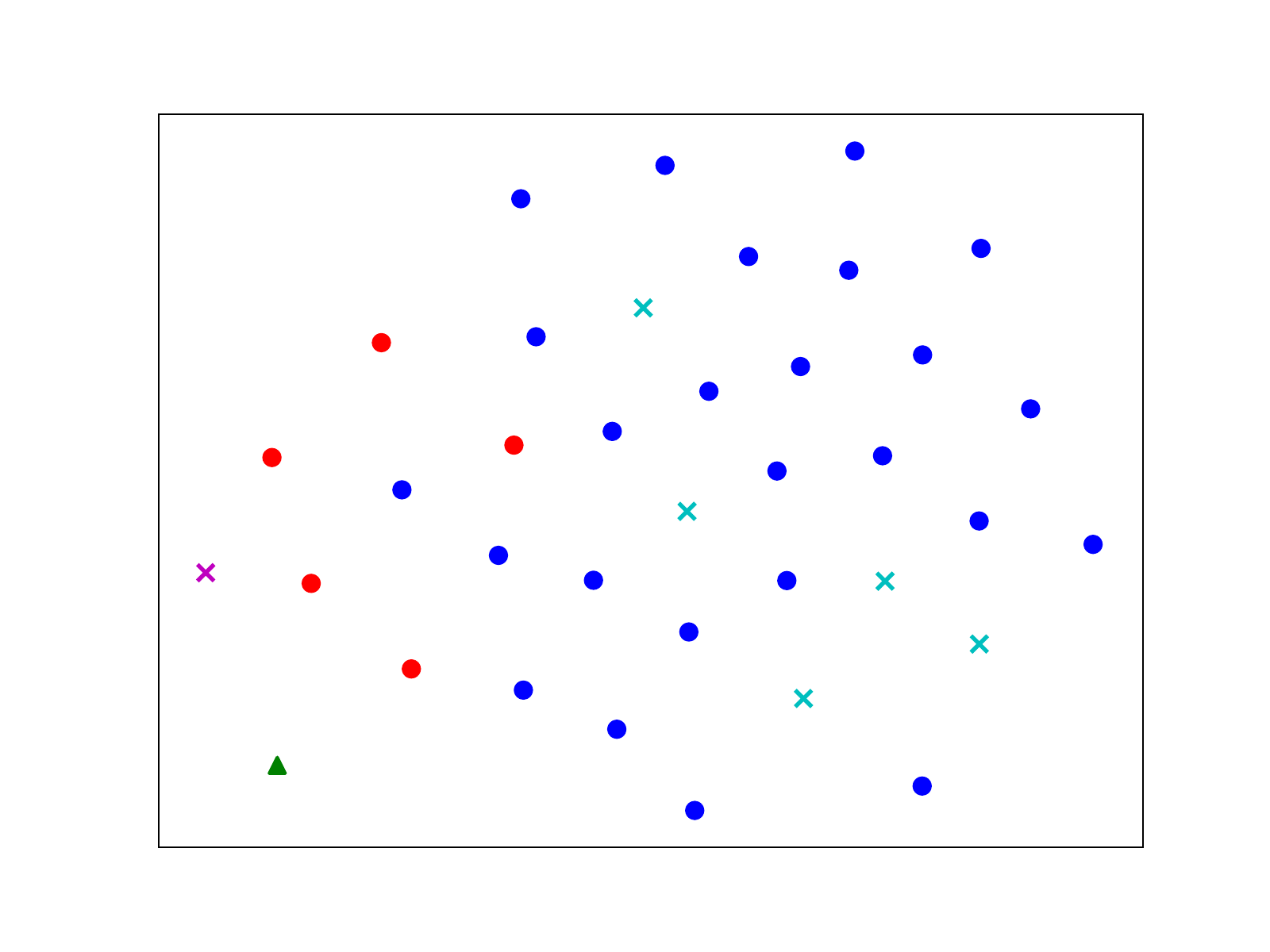}&
    \includegraphics[width=16em]{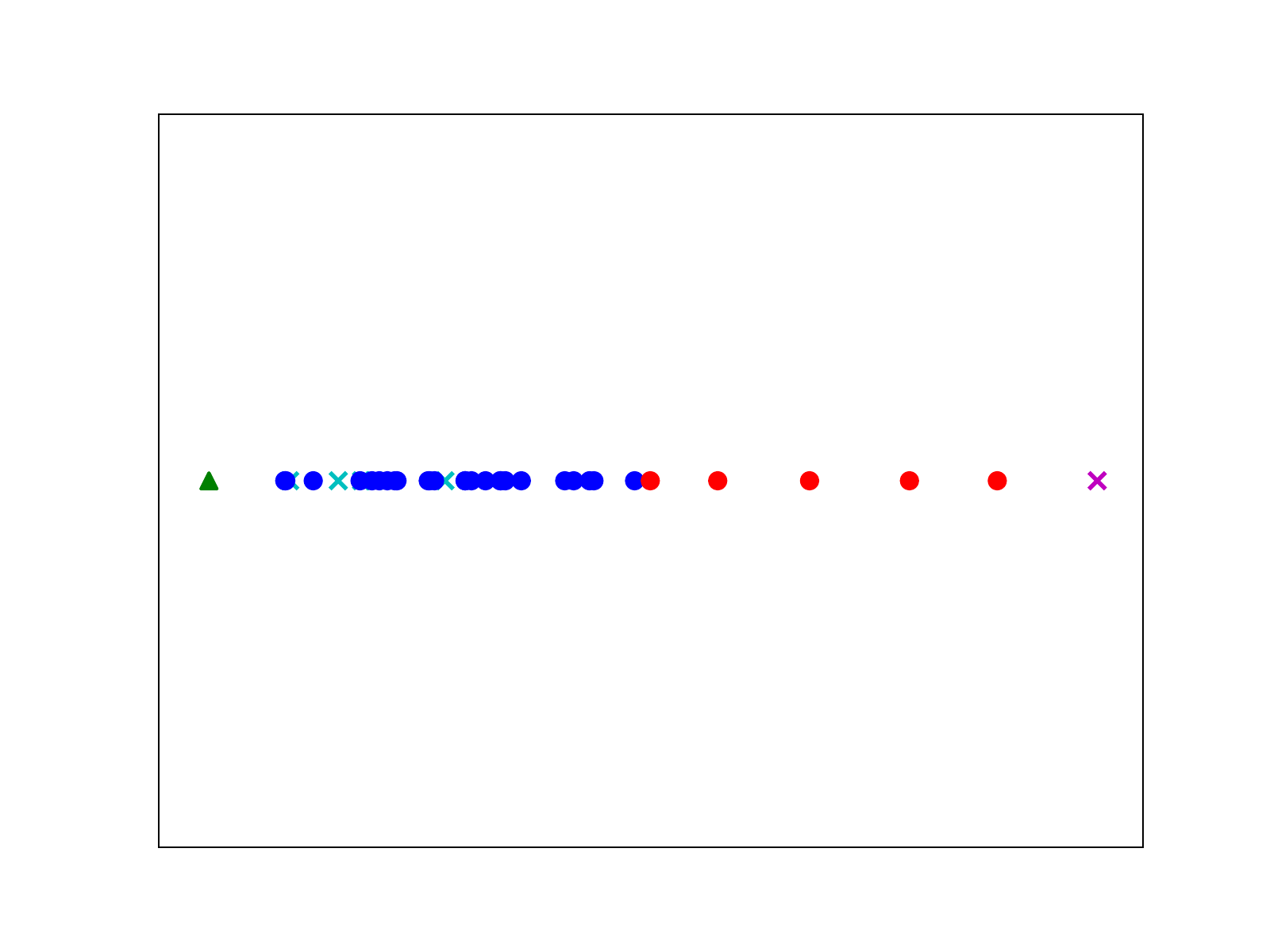}
    \\
    (a) Original space & (b) Embedded space    
    & (c) Linearly projected space \\
  \end{tabular}}
  \caption{Visualization of support and query instances in the a) original space $\vec{x}$,
    b) embedded space $\phi([\vec{x},\vec{r}])$,
    c) one-dimensional linearly projected space $\hat{\vec{w}}^{\top}\phi([\vec{x},\vec{r}])$
    on the WDBC dataset by the proposed method. A cyan `x' is a normal support instance, a magenta `x' is an anomalous support instance, a blue `o' is a normal query instance, a red `o' is an anomalous query instance, and a green `$\triangle$' is center. In (a) and (b), we used t-SNE~\cite{maaten2008visualizing} for visualizing in a two-dimensional space.}
  \label{fig:viz}
\end{figure*}

Figure~\ref{fig:viz} shows
an example of visualization of support and query instances in the original,
embedded and linearly projected spaces on the WDBC dataset by the proposed method.
With the original space (a),
an anomalous query instance (red `o') was located further from the anomalous support instance (magenta `x')
than some normal query instances (blue `o').
Therefore, the anomalous query instance could not be detected as anomaly in the original space.
By embedding (b), anomalous instances were located more closely together than the original space.
However, there was the center (green '$\triangle$') near anomalous instances,
and there were some normal instances in the region of anomalous instances.
By linear projection (c), anomalous instances (magenta `x' and red `o') were located
further from the center than normal instances (cyan `x' and blue `o'),
where the anomalous instances were appropriately detected by the distance from the center.

Table~\ref{tab:ablation} shows the average test AUC on target tasks on ablation study.
WoNN was the proposed method without neural networks,
where original attribute vectors were linearly projected to a one-dimensional space,
$\hat{\vec{w}}^{\top}\vec{x}$,
where $\hat{\vec{w}}$ was determined by the generalized eigenvalue problem.
WoProj was the proposed method without the linear projection,
where anomaly scores were calculated by the distance
from the center before the linear projection
$a(\vec{x}|\mathcal{S})=\parallel\phi([\vec{x},\vec{r}])-\vec{c}\parallel^{2}$.
WoAnomaly was the proposed method without anomalous instances,
which was the method described in Section~\ref{sec:normal}.
The proposed method was better than WoNN and WoProj.
This result indicates the effectiveness of the neural networks
and linear projection in our model.
The performance of WoAnomaly was
better than existing unsupervised methods in Table~\ref{tab:auc}, SVDD, OSVM, IF, and LOF.
This result demonstrates the effectiveness of the proposed method without anomaly
described in Section~\ref{sec:normal}.
The proposed method with anomaly improved the performance from that without anomaly
by effectively using anomaly information via the neural networks
with the generalized eigenvalue problem solver layer.

\begin{table}[t!]
  \centering
  \caption{Average test AUC on target tasks on ablation study of the proposed method by WoNN (without neural networks), WoProj (without linear projection), and WoAnomaly (without support anomalous instances).}
  \label{tab:ablation}
          \begin{tabular}{rrrrr}
      \hline
      Ours & WoNN & WoProj & WoAnomaly\\ 
      \hline
      0.913 & 0.880 & 0.874 & 0.853\\
      \hline
    \end{tabular}
  \end{table}

\begin{table}[t!]
  \centering
  \caption{Average computational time in seconds for training by
    the proposed method, SVDD, SSVDD, Proto, R2D2, and MAML.}
  \label{tab:time}
          \begin{tabular}{rrrrrr}
        \hline
          Ours & {\small SVDD} & {\small SSVDD} & {\small Proto} & {\small R2D2} & {\small MAML} \\
          \hline
        1,527 & 348 & 351 & 570 & 488 & 33,528 \\
        \hline
      \end{tabular}
\end{table}

\begin{figure}
  \centering
  \includegraphics[width=20em]{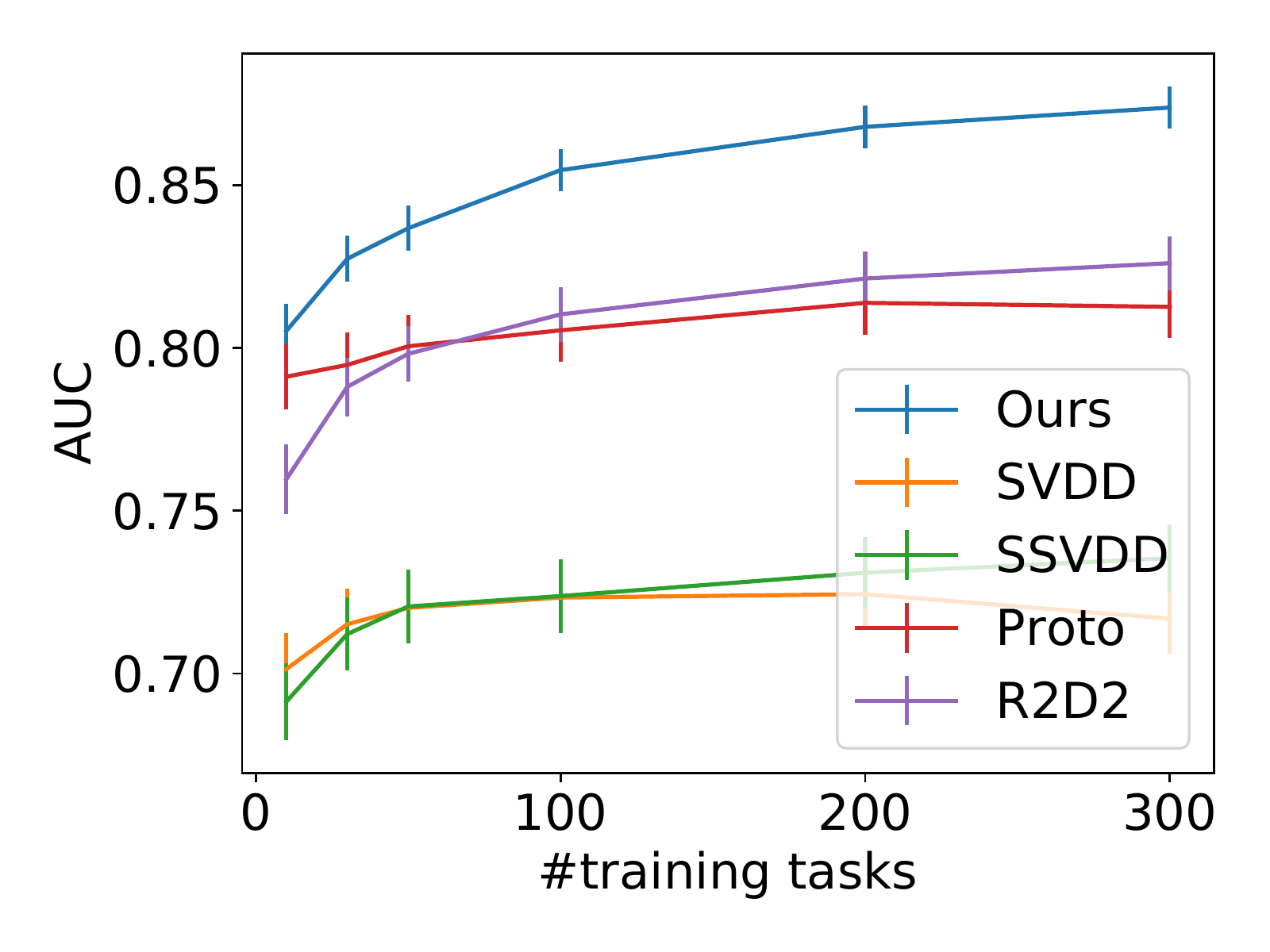}
  \caption{Average test AUC on target tasks
    with different numbers of training tasks
    on 22 outlier detection datasets.}
  \label{fig:auc_ndataset}
\end{figure}

Figure~\ref{fig:auc_ndataset}
shows the averaged test AUC on target tasks
with different numbers of training tasks
on 22 outlier detection datasets
by the proposed method, SVDD, SSVDD, Proto, and R2D2.
All the methods improved the performance as the number of training tasks increased.
The proposed method achieved the best test AUC with all cases.

Table~\ref{tab:time}
shows the average computational time in seconds for training by
the proposed method, SVDD, SSVDD, Proto, R2D2, and MAML,
on computers with a 2.60GHz CPU.
The proposed method had longer training time than SVDD, SVDD, Proto, and R2D2,
since the proposed method calculated the task representation and linear projection
for each task.
However, it was much faster than MAML,
which required the gradient of multiple gradient steps for each task.

\section{Conclusion}
\label{sec:conclusion}

We proposed a neural network-based meta-learning method for anomaly detection,
where our model is trained using multiple datasets
to improve the performance on unseen tasks.
We experimentally confirmed the effectiveness of the proposed method compared with
existing anomaly detection and few-shot learning methods.
For future work, we would like to extend the proposed method
for semi-supervised settings~\cite{blanchard2010semi},
where unlabeled instances, labeled normal and labeled anomalous instances are given as support sets.

\bibliographystyle{abbrv}
\bibliography{icml2021anomaly}

\appendix
\section{Derivation of Eq.~(3)}
\label{app:eq3}

\begin{align}
  \lefteqn{\frac{1}{N_{\mathrm{A}}}\sum_{\vec{x}\in\mathcal{S}_{\mathrm{A}}}a(\vec{x}|\mathcal{S})}
  \nonumber\\
  &=\frac{1}{N_{\mathrm{A}}}\sum_{\vec{x}\in\mathcal{S}_{\mathrm{A}}}\parallel\hat{\vec{w}}^{\top}\phi([\vec{x},\vec{r}])-\hat{\vec{w}}^{\top}\vec{c}\parallel^{2}\nonumber\\
  &=\frac{1}{N_{\mathrm{A}}}\sum_{\vec{x}\in\mathcal{S}_{\mathrm{A}}}(\hat{\vec{w}}^{\top}(\phi([\vec{x},\vec{r}])-\vec{c}))^{\top}(\hat{\vec{w}}^{\top}(\phi([\vec{x},\vec{r}])-\vec{c}))\nonumber\\
  &=\frac{1}{N_{\mathrm{A}}}\sum_{\vec{x}\in\mathcal{S}_{\mathrm{A}}}\hat{\vec{w}}^{\top}(\phi([\vec{x},\vec{r}])-\vec{c})(\phi([\vec{x},\vec{r}])-\vec{c})^{\top}\hat{\vec{w}}\nonumber\\
  &=
  \hat{\vec{w}}^{\top}\frac{1}{N_{\mathrm{A}}}\sum_{\vec{x}\in\mathcal{S}_{\mathrm{A}}}(\phi([\vec{x},\vec{r}])-\vec{c})(\phi([\vec{x},\vec{r}])-\vec{c})^{\top}\hat{\vec{w}}\nonumber\\
  &=
  \hat{\vec{w}}^{\top}\vec{S}_{\mathrm{A}}\hat{\vec{w}}
\end{align}

Similarly,
\begin{align}
  \frac{1}{N_{\mathrm{N}}}\sum_{\vec{x}\in\mathcal{S}_{\mathrm{N}}}a(\vec{x}|\mathcal{S})+\eta\parallel\vec{w}\parallel^{2}=
  \hat{\vec{w}}^{\top}\vec{S}_{\mathrm{N}}\hat{\vec{w}}
\end{align}

\section{Derivation of Eq.~(7)}
\label{app:eq7}

When the number of anomalous instance is one,
$\vec{S}_{\mathrm{A}}\hat{\vec{w}}$ is on the same direction with $\phi([\vec{x},\vec{r}])-\vec{c}$ as follows,
\begin{align}
  \vec{S}_{\mathrm{A}}\hat{\vec{w}}=(\phi([\vec{x},\vec{r}])-\vec{c})(\phi([\vec{x},\vec{r}])-\vec{c})^{\top}\hat{\vec{w}}
  \propto \phi([\vec{x},\vec{r}])-\vec{c}.
\end{align}
Since $\vec{S}_{\mathrm{A}}\hat{\vec{w}}=\lambda\vec{S}_{\mathrm{N}}\hat{\vec{w}}$,
we obtain
\begin{align}
  \phi([\vec{x},\vec{r}])-\vec{c}\propto\lambda\vec{S}_{\mathrm{N}}\hat{\vec{w}},
\end{align}
and then
\begin{align}
\hat{\vec{w}} \propto \vec{S}_{\mathrm{N}}^{-1}(\phi([\vec{x},\vec{r}])-\vec{c}).
\end{align}

\section{Data}
\label{app:data}

Table~\ref{tab:data} shows the statistics for each dataset used in our experiments.

\begin{table*}[t!]
  \centering
  \caption{Statistics of datasets used in our experiments: The number of training, validation and target tasks, number of normal and anomalous instances, and number of attribute for each dataset.}
  \label{tab:data}
  \begin{tabular}{lrrrrrr}
    \hline
    Data & Train & Valid & Target & Normal & Anomaly & Attribute\\
\hline
ALOI & 400 & 50 & 50 & 48492 & 1508 & 27\\
Annthyroid & 400 & 50 & 50 & 6666 & 350 & 21\\
Arrhythmia & 400 & 50 & 50 & 244 & 61 & 259\\
Cardiotocography & 400 & 50 & 50 & 1655 & 413 & 21\\
Glass & 400 & 50 & 50 & 205 & 9 & 7\\
HeartDisease & 400 & 50 & 50 & 150 & 37 & 13\\
Hepatitis & 400 & 50 & 50 & 67 & 7 & 19\\
InternetAds & 400 & 50 & 50 & 1598 & 177 & 1555\\
Ionosphere & 400 & 50 & 50 & 225 & 126 & 32\\
KDDCup99 & 400 & 50 & 50 & 60593 & 246 & 79\\
Lymphography & 400 & 50 & 50 & 142 & 6 & 47\\
PageBlocks & 400 & 50 & 50 & 4913 & 258 & 10\\
Parkinson & 400 & 50 & 50 & 48 & 12 & 22\\
PenDigits & 400 & 50 & 50 & 9848 & 20 & 16\\
Pima & 400 & 50 & 50 & 500 & 125 & 8\\
Shuttle & 400 & 50 & 50 & 1000 & 13 & 9\\
SpamBase & 400 & 50 & 50 & 2788 & 697 & 57\\
Stamps & 400 & 50 & 50 & 309 & 16 & 9\\
WBC & 400 & 50 & 50 & 444 & 10 & 9\\
WDBC & 400 & 50 & 50 & 357 & 10 & 30\\
Waveform & 400 & 50 & 50 & 3343 & 100 & 21\\
Wilt & 400 & 50 & 50 & 4578 & 93 & 5\\
Landminedata & 23 & 2 & 4 & 651 & 38 & 9\\
Doorbell & 66 & 4 & 10 & 200 & 50 & 115\\
Thermostat & 64 & 6 & 10 & 200 & 50 & 115\\
Doorbell & 60 & 15 & 5 & 200 & 50 & 115\\
Monitor & 61 & 9 & 10 & 200 & 50 & 115\\
Camera & 63 & 7 & 10 & 200 & 50 & 115\\
Camera & 64 & 6 & 10 & 200 & 50 & 115\\
Webcam & 68 & 7 & 5 & 200 & 50 & 115\\
Camera & 61 & 9 & 10 & 200 & 50 & 115\\
Camera & 61 & 9 & 10 & 200 & 50 & 115\\
\hline
  \end{tabular}
\end{table*}

\end{document}